\documentclass[runningheads]{llncs}
\usepackage{graphicx}
\usepackage{amsmath,amssymb} 
\usepackage{times}
\usepackage{epsfig}
\usepackage{amsmath}
\usepackage{multirow}
\usepackage{comment}
\usepackage{subfigure}
\usepackage{enumitem}
\usepackage{color}
\usepackage[width=122mm,left=12mm,paperwidth=146mm,height=193mm,top=12mm,paperheight=217mm]{geometry}
\newcommand*\samethanks[1][\value{footnote}]{\footnotemark[#1]}
\begin{document}
	\pagestyle{headings}
	\mainmatter

	\title{Online Human Action Detection using Joint Classification-Regression Recurrent Neural Networks} 
	
	\titlerunning{Online Human Action Detection using Joint Classification-Regression RNNs}
	
	\authorrunning{Li, Y.\and Lan, C.\and Xing, J.\and Zeng, W.\and Yuan C.\and Liu J.}
	
	\author{Yanghao Li$^1$ \and Cuiling Lan$^2$\thanks{Corresponding author. This work was done at Microsoft Research Asia.} \and Junliang Xing$^3$ \and Wenjun Zeng$^2$ \and\\ Chunfeng Yuan$^3$ \and Jiaying Liu$^{1}$\samethanks}
	\institute{$^1$ Institute of Computer Science and Technology, Peking University\\
	$^2$ Microsoft Research Asia ~~
	$^3$ Institute of Automation, Chinese Academy of Sciences\\
	\email{\{lyttonhao, liujiaying\}@pku.edu.cn},~~\\ \email{\{culan, wezeng\}@microsoft.com},~~\email{\{jlxing, cfyuan\}@nlpr.ia.ac.cn}}
\maketitle


\begin{abstract}
Human action recognition from well-segmented 3D skeleton data has been intensively studied and has been attracting an increasing attention. Online action detection goes one step further and is more challenging, which identifies the action type and localizes the action positions on the fly from the untrimmed stream data. In this paper, we study the problem of online action detection from streaming skeleton data. We propose a multi-task end-to-end Joint Classification-Regression Recurrent Neural Network to better explore the action type and temporal localization information. By employing a joint classification and regression optimization objective, this network is capable of automatically localizing the start and end points of actions more accurately. Specifically, by leveraging the merits of the deep Long Short-Term Memory (LSTM) subnetwork, the proposed model automatically captures the complex long-range temporal dynamics, which naturally avoids the typical sliding window design and thus ensures high computational efficiency. Furthermore, the subtask of regression optimization provides the ability to forecast the action prior to its occurrence. To evaluate our proposed model, we build a large streaming video dataset with annotations. Experimental results on our dataset and the public G3D dataset both demonstrate very promising performance of our scheme.

\end{abstract}

\begin{section}{Introduction}

Human action detection is an important problem in computer vision, which has broad practical applications like visual surveillance, human-computer interaction and intelligent robot navigation. Unlike action recognition and offline action detection, which determine the action after it is fully observed, online action detection aims to detect the action on the fly, as early as possible. It is much desirable to accurately and timely localize the start point and end point of an action along the time and determine the action type as illustrated in Fig. \ref{fig:example}. Besides, it is also desirable to forecast the start and end of the actions prior to their occurrence. For example, for intelligent robot system, in addition to the accurate detection of actions, it would also be appreciated if it can predict the start of the impending action or the end of the ongoing actions and then get something ready for the person it serves, e.g., passing towels when he/she finishes washing hands. Therefore, the detection and forecast system could respond to impending or ongoing events accurately and as soon as possible, to provide better user experiences.

For human action recognition and detection, many research works have been designed for RGB videos recorded by 2D cameras in the past couple of decades \cite{CVIU11SurveyAction}. In recent years, with the prevalence of the affordable color-depth sensing cameras, such as the Microsoft Kinect \cite{Kinect}, it is much easier and cheaper to obtain depth data and thus the 3D skeleton of human body (see skeleton examples in Fig. \ref{fig:example}). Biological observations suggest that skeleton, as an intrinsic high level representation, is very valuable information for recognizing actions by humans \cite{PP73Perception}. 
In comparison to RGB video, such high level human representation by skeleton is robust to illumination and clustered background \cite{SkeletonReview16}, but may not be appropriate for recognizing fine-grained actions with marginal differences. Taking the advantages of skeleton representation, in this paper, we investigate skeleton based human action detection. The addition of RGB information may result in better performance and will be addressed in the future work.   


\begin{figure}[t]
  \centerline{\includegraphics[width=0.62\linewidth]{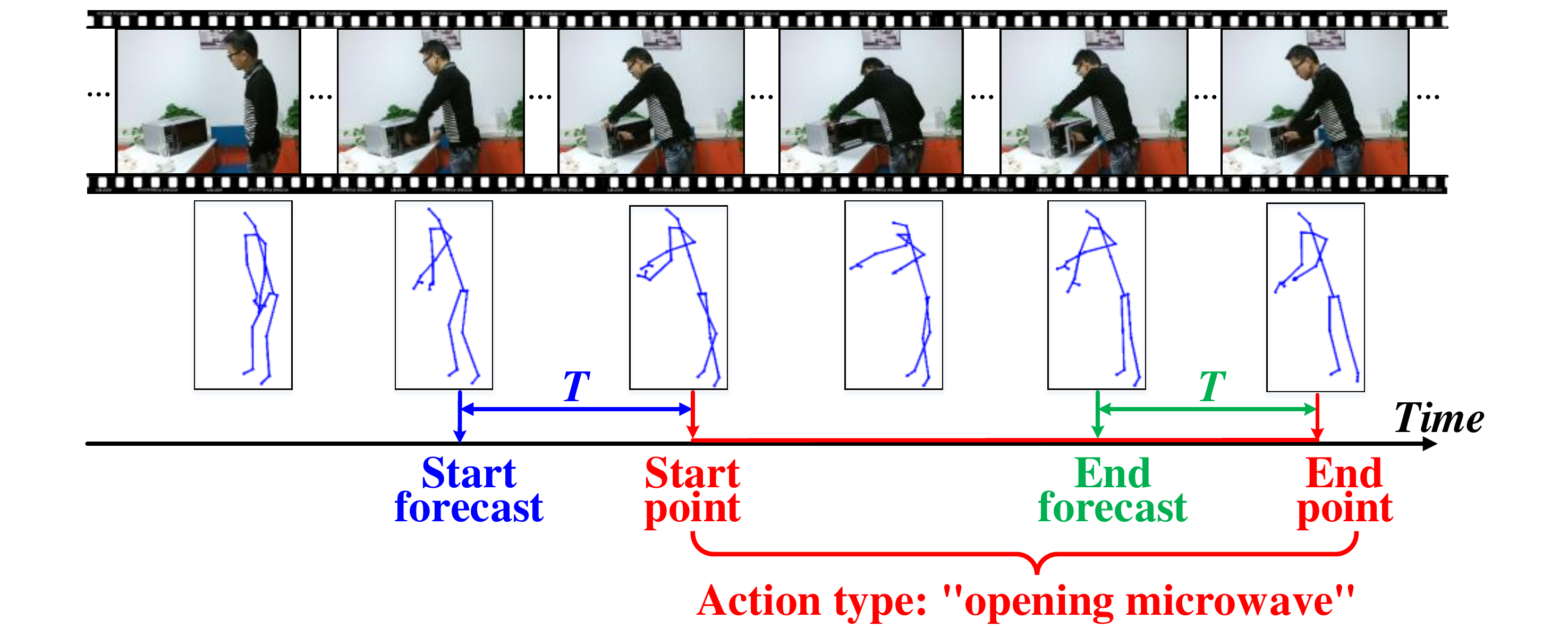}}
\caption{Illustration of online action detection. It aims to determine the action type and the localization on the fly. It is also desirable to forecast the start and end points (e.g., $T$ frames ahead).}
\label{fig:example}
\end{figure}

Although online action detection is of great importance, there are very few works specially designed for it \cite{hoai2014max,zanfir2013moving}. Moreover, efficient exploitation of the advanced recurrent neural network (RNN) has not been well studied for the efficient temporal localization of actions. Most published methods are designed for offline detection \cite{Lear2014}, which performs detection after fully observing the sequence. To localize the action, most of previous works employ a sliding window design \cite{siva2011weakly,hoai2014max,motionappearance14,sharaf2015real}, which divides the sequence into overlapped clips before action recognition/classification is performed on each clip. Such sliding window design has low computational efficiency. A method that divides the continuous sequence into short clips with the shot boundary detected by computing the color histogram and motion histogram was proposed in \cite{CUHKSIATTHUMOS15}. However, indirect modeling of action localization in such an unsupervised manner does not provide satisfactory performance. An algorithm which can intelligently localize the actions on the fly is much expected, being suitable for the streaming sequence with actions of uncertain length. For action recognition on a segmented clip, deep learning methods, such as convolutional neural networks and recurrent neural networks, have been shown to have superior performances on feature representation and temporal dynamics modeling \cite{Wu2015ModelingTem,donahue2015long,du2015hierarchical,Zhu2016Co}. However, how to design an efficient online action detection system that leverages the neural network for the untrimmed streaming data is not well studied.


\begin{figure}[t] 
	\centerline{\includegraphics[width=0.62\linewidth]{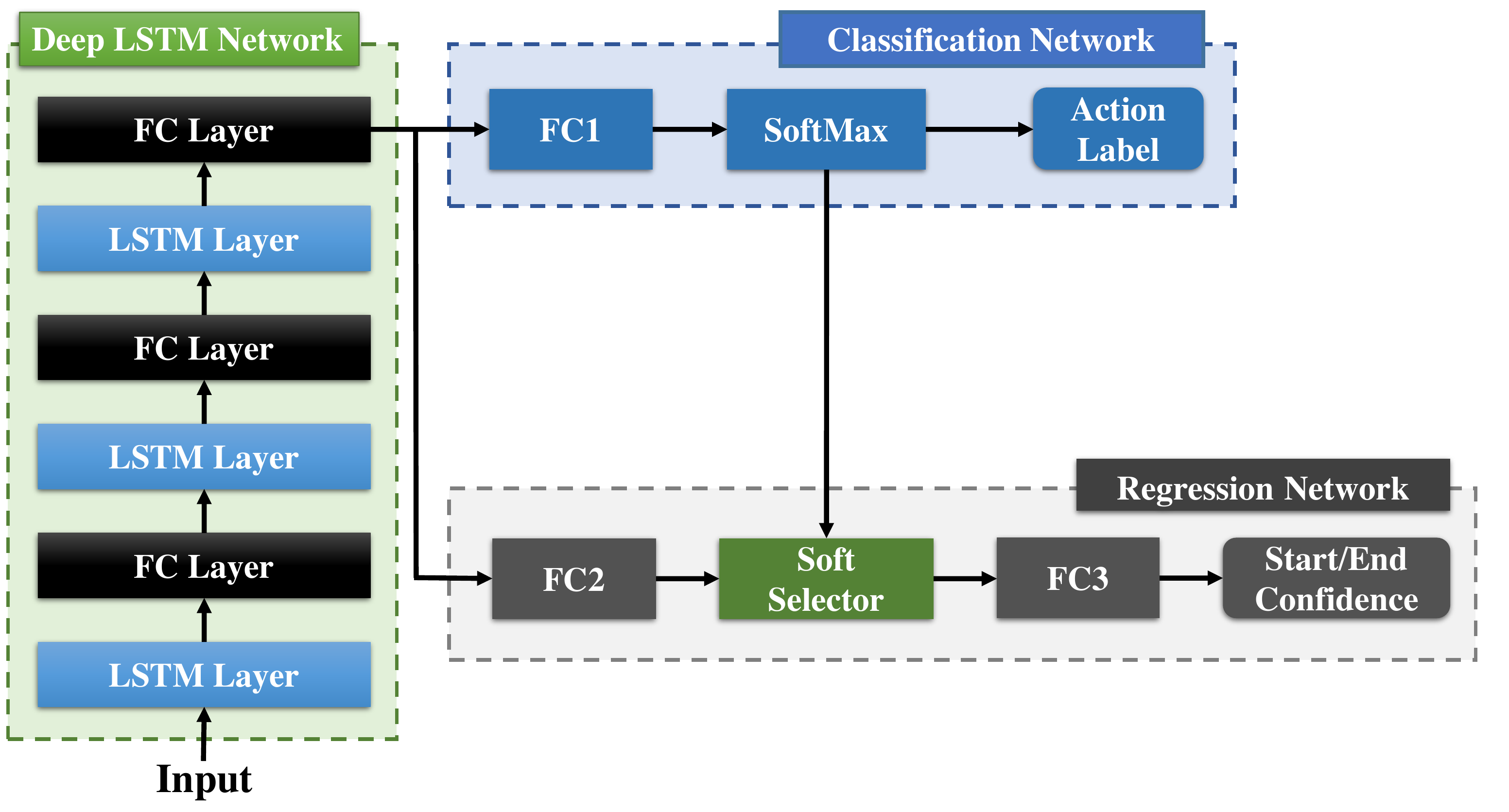}}
	\caption{Architecture of the proposed joint classification-regression RNN framework for online action detection and forecasting.}
	\label{fig:flowchart}
\end{figure}

In this paper, we propose a Joint Classification-Regression Recurrent Neural Network to accurately detect the actions and localize the start and end positions of the actions on the fly from the streaming data. Fig. \ref{fig:flowchart} shows the architecture of the proposed framework. Specifically, we use LSTM \cite{hochreiter1997long} as the recurrent layers to perform automatic feature learning and long-range temporal dynamics modeling. Our network is end-to-end trainable by optimizing a joint objective function of frame-wise action classification and temporal localization regression. On one hand, we perform frame-wise action classification, which aims to detect the actions timely. On the other hand, to better localize the start and end of actions, we incorporate the regression of the start and end points of actions into the network. We can forecast their occurrences in advance based on the regressed curve. We train this classification and regression network jointly to obtain high detection accuracy. Note that the detection is performed frame-by-frame and the temporal information is automatically learnt by the deep LSTM network without requiring a sliding window design, which is time efficient.

The main contributions of this paper are summarized as follows:
\begin{itemize}
\item We investigate the new problem of online action detection for streaming skeleton data by leveraging recurrent neural network.
\item We propose an end-to-end Joint Classification-Regression RNN to address our target problem. 
{Our method leverages the advantages of RNNs for frame-wise action detection and forecasting without requiring a sliding window design and explicit looking forward or backward.}
\item We build a large action dataset for the task of online action detection from streaming sequence.
\end{itemize}
\end{section}

\begin{section}{Related Work}\label{sec:related}
\begin{subsection}{Action Recognition and Action Detection}

Action recognition and detection have attracted a lot of research interests in recent years. Most methods are designed for action recognition \cite{donahue2015long,du2015hierarchical,simonyan2014two}, i.e., to recognize the action type from a well-segmented sequence, or offline action detection \cite{wei2013concurrent,sharaf2015real,siva2011weakly,tian2013spatiotemporal}. However, in many applications it is desirable to recognize the action on the fly, without waiting for the completion of the action, e.g., in human computer interaction to reduce the response delay. In \cite{hoai2014max}, a learning formulation based on a structural SVM is proposed to recognize partial events, enabling early detection. {To reduce the observational latency of human action recognition, a non-parametric moving pose framework \cite{zanfir2013moving} and a dynamic integral bag-of-words approach \cite{ryoo2011human} are proposed respectively to detect actions earlier.} Our model goes beyond early detection. Besides providing frame-wise class information, it forecasts the occurrence of start and end of actions.

To localize actions in streaming video sequence, existing detection methods utilize either sliding-window scheme \cite{siva2011weakly,hoai2014max,motionappearance14,sharaf2015real}, or action proposal approaches \cite{jain2014action,yu2015fast,CUHKSIATTHUMOS15}. These methods usually have low computational efficiency or unsatisfactory localization accuracy due to the overlapping design and unsupervised localization approach. Besides, it is not easy to determine the sliding-window size.

Our framework aims to address the online action detection in such a way that it can predict the action at each time slot efficiently without requiring a sliding window design. We use the regression design to determine the start/end points learned in a supervised manner during the training, enabling the localization being more accurate. Furthermore, it forecasts the start of the impending or end of the ongoing actions.
\end{subsection}

\begin{subsection}{Deep Learning} 
Recently, deep learning has been exploited for action recognition \cite{simonyan2014two}. Instead of using hand-crafted features, deep learning can automatically learn robust feature representations from raw data. To model temporal dynamics, RNNs have also been used for action recognition. A Long-term Recurrent Convolutional Network (LRCN) \cite{donahue2015long} is proposed for activity recognition, where the LRCN model contains several Convolutional Neural Network (CNN) layers to extract visual features followed by LSTM layers to handle temporal dynamics. A hybrid deep learning framework is proposed for video classification \cite{Wu2015ModelingTem}, where LSTM networks are applied on top of the two types of CNN-based features related to the spatial and the short-term motion information. For skeleton data, hierarchical RNN \cite{du2015hierarchical} and fully connected LSTM \cite{Zhu2016Co} are investigated to model the temporal dynamics for skeleton based action recognition.


Despite a lot of efforts in action recognition, which uses pre-segmented sequences, there are few works on applying RNNs for the action detection and forecasting tasks. Motivated by the advantages of RNNs in sequence learning and some other online detection tasks (e.g. audio onset \cite{bock2012online} and driver distraction \cite{wollmer2011online} detection), 
we propose a Joint Classification and Regression RNN to automatically localize the action location and determine the action type on the fly. In our framework, the designed LSTM network simultaneously plays the role of feature extraction and temporal dynamic modeling. Thanks to the long-short term memorizing function of LSTM, we do not need to assign an observation window as in the sliding window based approaches for the action type determination and avoid the repeat calculation. This enables our design to have superior detection performance with low computation complexity.

\end{subsection}
\end{section}

\begin{section}{Problem Formulation}\label{sec:problem}
In this section, we formulate the online action detection problem. To help clarify the differences, offline action detection is first discussed. 
\begin{subsection}{Offline Action Detection}
Given a video observation $V = \{v_{0}, ..., v_{N-1}\}$ composed of frames from time $0$ to $N-1$, the goal of action detection is to determine whether a frame $v_t$ at time $t$ belongs to an action among the predefined $M$ action classes.


Without loss of generality, the target classes for the frame $v_t$ are denoted by a label vector ${\mathbf{y}}_{t} \in R^{1 \times (M+1)}$, where $y_{t,j} = 1$ means the presence of an action of class $j$ at this frame and $y_{t,j} = 0$ means absence of this action. Besides the $M$ classes of actions, a blank class is added to represent the situation in which the current frame does not belong to any predefined actions. Since the entire sequence is known, the determination of the classes at each time slot is to maximize the posterior probability
\begin{equation}\label{eq:action_detection}
\mathbf{y}_t^{*}=\operatornamewithlimits{argmax}_{\mathbf{y}_t}{P(\mathbf{y}_t | V)},
\end{equation}
where $\mathbf{y}_t$ is the possible action label vector for frame $v_t$. Therefore, conditioned on the entire sequence $V$, the action label with the maximum probability $P(\mathbf{y}_t | V)$ is chosen to be the status of frame $v_t$ in the sequence.

{According to the action label of each frame, an occurring action $i$ can be represented in the form $d_i = \{g_i, t_{i,start}, t_{i,end}\}$, where $g_i$ denotes the class type of the action $i$, $t_{i,start}$ and $t_{i,end}$ correspond to the starting and ending time of the action, respectively.}

\end{subsection}

\begin{subsection}{Online Action Detection}

In contrast to offline action detection, which makes use of the whole video to make decisions, online detection is required to determine which actions the current frame belongs to without using future information. Thus, the method must automatically estimate the start time and status of the current action. The problem can be formulated as
\begin{equation}\label{eq:online_detection}
\mathbf{y}_t^{*}=\operatornamewithlimits{argmax}_{\mathbf{y}_t}{P(\mathbf{y}_t | {v_0, ..., v_t})}.
\end{equation}

Besides determining the action label, an online action detection system for streaming data is also expected to predict the starting and ending time of an action. We should be aware of the occurrence of the action as early as possible and be able to predict the end of the action. For example, for an action $d_i = \{g_i, t_{i,start}, t_{i,end}\}$, the system is expected to forecast the start and end of the action during $[t_{i,start}-T, t_{i,start}]$ and $[t_{i,end}-T, t_{i,end}]$, respectively, ahead its occurrence. $T$ could be considered as the expected forecasting time in statistic. We define the optimization problem as

\begin{equation}
(\mathbf{y}_t^{*}, \mathbf{a}_t^{*}, \mathbf{b}_t^{*}) =\operatornamewithlimits{argmax}_{\mathbf{y}_t,\mathbf{a}_t,\mathbf{b}_t}{P({\bf{y}}_t, \mathbf{a}_t, \mathbf{b}_t | {v_0, ..., v_t})},
\end{equation}
where $\mathbf{a}_t$ and $\mathbf{b}_t$ are two vectors, denoting whether actions are to start or to stop within the following $T$ frames, respectively. For example, $a_{t,g_i} = 1$ means the action of class $g_i$ will start within $T$ frames.

\end{subsection}

\end{section}

\begin{section}{Joint Classification-Regression RNN for Online Action Detection}\label{sec:proposed}

We propose an end-to-end Joint Classification-Regression framework based on RNN to address the online action detection problem. Fig. \ref{fig:flowchart} shows the architecture of the proposed network, which has a shared deep LSTM network for feature extraction and temporal dynamic modeling, a classification subnetwork and a regression subnetwork. Note that we construct the deep LSTM network by stacking three LSTM layers and three non-linear fully-connected (FC) layers to have powerful learning capability. We first train the classification network for the frame-wise action classification. Then under the guidance of the classification results through the Soft Selector, we train the regressor to obtain more accurate localization of the start and end time points. 

In the following, we first briefly review the RNNs and LSTM to make the paper self-contained. Then we introduce our proposed  joint classification-regression network for online action detection.

\begin{subsection}{Overview of RNN and LSTM}

In contrast to traditional feedforward neural networks, RNNs have self-connected recurrent connections which model the temporal evolution. The output response $\mathbf{h}_t$ of a recurrent hidden layer can be formulated as follows \cite{graves2012supervised}
\begin{equation}
\mathbf{h}_t = \theta_h (\mathbf{W}_{xh}\mathbf{x}_t + \mathbf{W}_{hh}\mathbf{h}_{t-1} + \mathbf{b}_h),
\end{equation}
where $\mathbf{W}_{xh}$ and $\mathbf{W}_{hh}$ are mapping matrices from the current inputs $\mathbf{x}_t$ to the hidden layer $h$ and the hidden layer to itself. $\mathbf{b}_h$ denotes the bias vector. $\theta_h$ is the activation function in the hidden layer.

\begin{figure}[t] 
	\centering
	\begin{minipage}[t]{0.40\linewidth}
		\centering\includegraphics[width=1\linewidth]{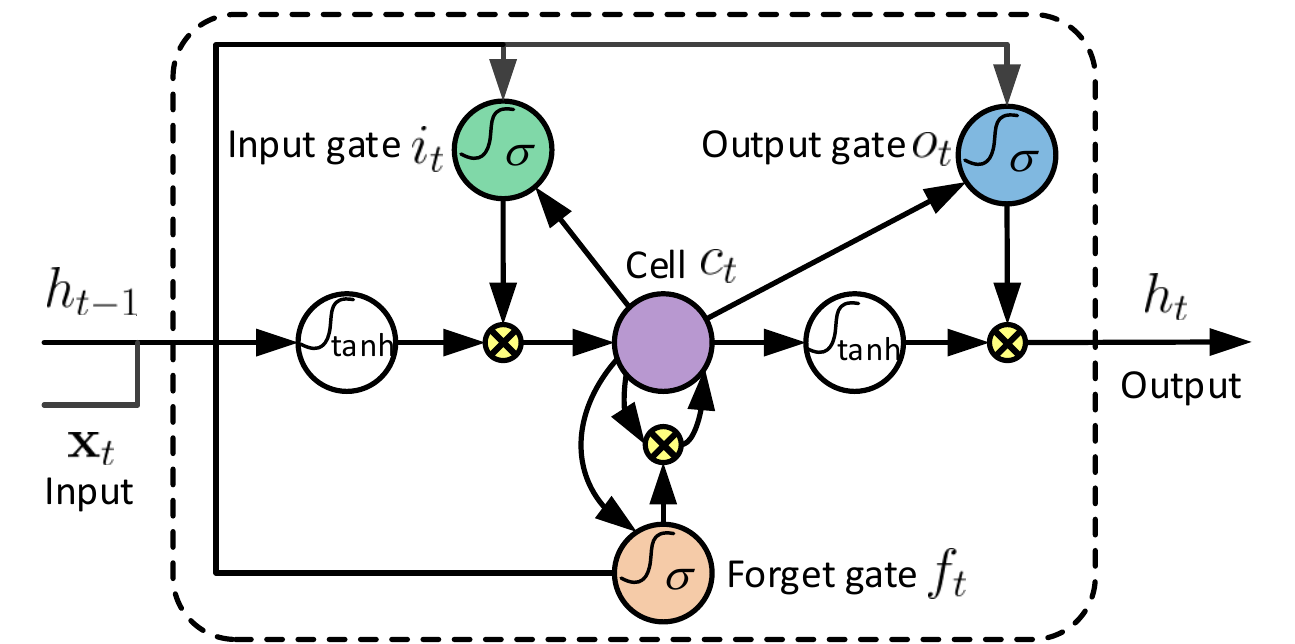}
		\caption{The structure of an LSTM neuron, which contains an input gate $i_t$, a forget gate $f_t$, and an output gate $i_t$. Information is saved in the cell $c_t$.}
		\label{fig:lstm}
	\end{minipage}
	\hspace{3ex}   
	\begin{minipage}[t]{0.45\linewidth}
		\centering{\includegraphics[width=1\linewidth]{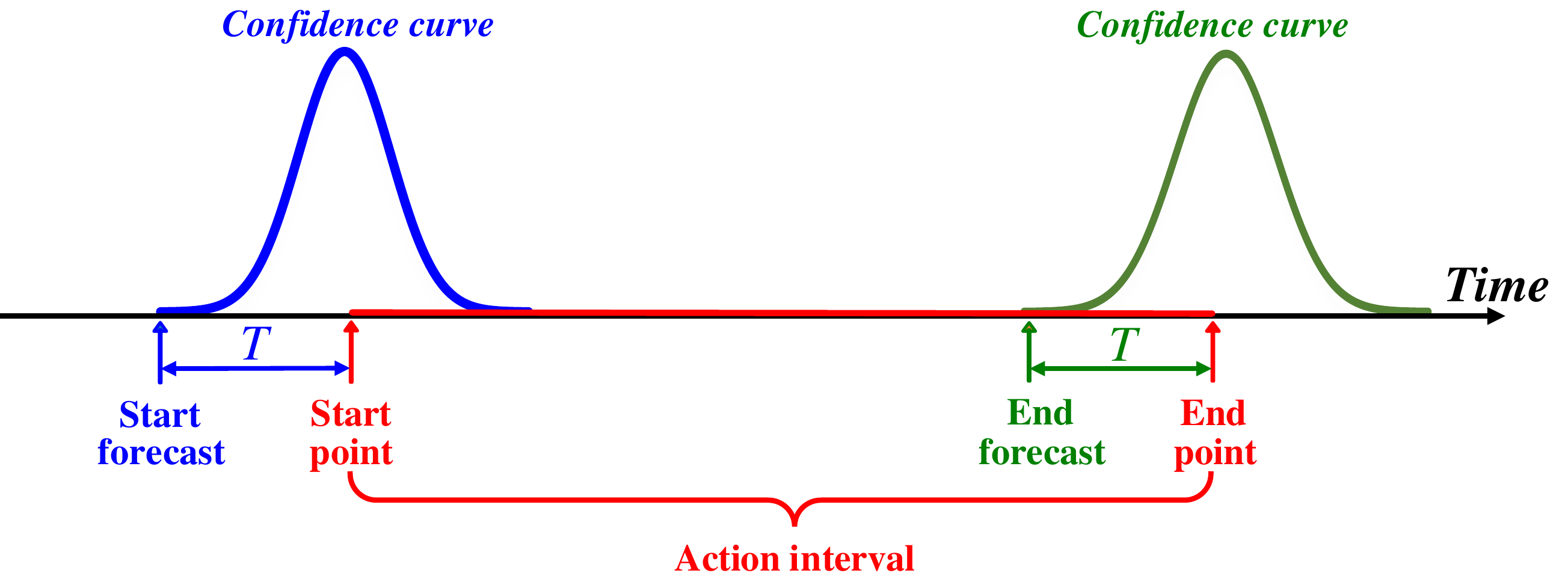}}
		\caption{Illustration of the confidence values around the start point and end point, which follow Gaussian-like curves with the confidence value 1 at the start and end point.}
		\label{fig:confidence}
	\end{minipage}
\end{figure}
%

The above RNNs have difficulty in learning long range dependencies \cite{Hochreiter2001gradientflow}, due to the vanishing gradient effect. To overcome this limitation, recurrent neural networks using LSTM \cite{hochreiter1997long,graves2012supervised,du2015hierarchical} has been designed to mitigate the vanishing gradient problem and learn the long-range contextual information of a temporal sequence. Fig. \ref{fig:lstm} illustrates a typical LSTM neuron. In addition to a hidden output $h_t$, an LSTM neuron contains an input gate $i_t$, a forget gate $f_t$, a memory cell $c_t$, and an output gate $o_t$. At each timestep, it can choose to read, write or reset the memory cell through the three gates. This strategy allows LSTM to memorize and access information many timesteps ago.


\end{subsection}

\begin{subsection}{Subnetwork for Classification Task}

We first train an end-to-end classification subnetwork for frame-wise action recognition. The structure of this classification subnetwork is shown in the upper part of Fig. \ref{fig:flowchart}. The frame first goes through the deep LSTM network, which is responsible for modeling the spatial structure and temporal dynamics. Then a fully-connected layer FC1 and a SoftMax layer are added for the classification of the current frame. The ouptput of the SoftMax layer is the probability distribution of the action classes ${\mathbf{y}}_t$. Following the problem formulation as described in Sec. \ref{sec:problem}, the objective function of this classification task is to minimize the cross-entropy loss function
\begin{equation}
\mathcal{L}_c(V) = - \frac{1}{N}\sum_{t = 0}^{N-1}\sum_{k=0}^{M}z_{t,k} \operatorname{ln} P(y_{t,k} | v_0,...v_t),
\end{equation}
where $z_{t,k}$ corresponds to the groundtruth label of frame $v_t$ for class $k$, $z_{t,k}=1$ means the groundtruth class is the $k^{th}$ class, $P(y_{t,k} | v_0,...v_t)$ denotes the estimated probability of being action classes $k$ of frame $v_t$. 

We train this network with Back Propagation Through Time (BPTT) \cite{werbos1990backpropagation} and use stochastic gradient descent with momentum to compute the derivatives of the objective function with respect to all parameters. To prevent over-fitting, we have utilized dropout at the three fully-connected layers.

\end{subsection}

\begin{subsection}{Joint Classification and Regression}

We fine-tune this network on the initialized classification model by jointly optimizing the classification and regression. Inspired by the Joint Classification-Regression models used in Random Forest \cite{glocker2012joint,schulter2014accurate} {for other tasks (e.g. segmentation \cite{glocker2012joint} and object detection \cite{schulter2014accurate})}, we propose our Joint learning to simultaneously make frame-wise classification, localize the start and end time points of actions, and to forecast them.

We define a confidence factor for each frame to measure the possibility of the current frame to be the start or end point of some action. To better localize the start or end point, we use a Gaussian-like curve to describe the confidences, which centralizes at the actual start (or end) point as illustrated in Fig. \ref{fig:confidence}. Taking the start point as an example, the confidence of the frame $v_t$ with respect to the start point of action $j$ is defined as
\begin{equation}
	c_{t}^s = e^{-{(t-s_j)^2}/{2\sigma^2}},
	\label{eq:gauss}
\end{equation}
where $s_j$ is the start point of the nearest (along time) action $j$ to the frame $v_t$, and $\sigma$ is the parameter which controls the shape of the confidence curve. Note that at the start point time, i.e., $t=s_j$, the confidence value is 1. Similarly, we denote the confidence of being the end point of one action as $c_{t}^e$. For the Gaussian-like curve, a lower confidence value suggests the current frame has larger distance from the start point and the peak point indicates the start point.

Such design has two benefits. First, it is easy to localize the start/end point by checking the regressed peak points. Second, this makes the designed system have the ability of forecasting. We can forecast the start (or end) of actions according to the current confidence response. We set a confidence threshold $\theta_s$ (or $\theta_e$) according to the sensitivity requirement of the system to predict the start (or end) point. When the current confidence value is larger than $\theta_s$ (or $\theta_e$), we consider that one action may start (or end) soon. Usually, larger threshold corresponds to a later response but a more accurate forecast.

Using the confidence as the target values, we include this regression problem as another task in our RNN model, as shown in the lower part of Fig. \ref{fig:flowchart}. This regression subnetwork consists of a non-linear fully-connected layer FC2, a Soft Selector layer, and a non-linear fully-connected layer FC3. Since we regress one type of confidence values for all the start points of different actions, we need to use the output of the action classification to guide the regression task. Therefore, we design a Soft Selector module to generate more specific features by fusing the output of SoftMax layer which
describes the probabilities of classification together with the output of the FC2 layer.

We achieve this by using class specific element-wise multiplication of the outputs of SoftMax and FC2 layer. The information from the SoftMax layer for the classification task plays the role of class-based feature selection over the output features of FC2 for the regression task. A simplified illustration about the Soft Selector model is shown in Fig. \ref{fig:selection}. Assume we have 5 action classes and the dimension of the FC2 layer output is reshaped to 7$\times$5. The vector (marked by circles) with the dimension of 5 from the SoftMax output denotes the probabilities of the current frame belonging to the 5 classes respectively. Element-wise multiplication is performed for each row of features and then integrating the SoftMax output plays the role of feature selection for different classes.

The final objective function of the Joint Classification-Regression is formulated as
\begin{eqnarray}
\vspace{-5mm}
\begin{aligned}
\mathcal{L}(V) & = \mathcal{L}_c(V) +\lambda\mathcal{L}_r(V) \\
&= -\frac{1}{N}  \sum_{t = 0}^{N-1} \Bigg[\bigg(\sum_{k=0}^{M}z_{t,k} \operatorname{ln}P(y_{t,k} | v_0,...v_t) \bigg) \Bigg.\\
& + \Bigg. \lambda\cdot\bigg(\ell(c_{t}^s, p_{t}^s) + \ell(c_{t}^e,p_{t}^e)\bigg)\Bigg],
\label{eq:final_loss}
\end{aligned}	
\vspace{-5mm}
\end{eqnarray}
where $p_{t}^s$ and $p_{t}^e$ are the predicted confidence values as start and end points, $\lambda$ is the weight for the regression task, $\ell$ is the regression loss function, which is defined as $\ell(x,y) = (x-y)^2$. In the training, the overall loss is a summarization of the loss from each frame $v_t$, where $0 \leq t < N$. For a frame $v_t$, its loss consists of the classification loss represented by the cross-entropy for the $M+1$ classes and the regression loss for identifying the start and end of the nearest action.

\begin{figure}[t]
	\centering\includegraphics[width=0.55\linewidth]{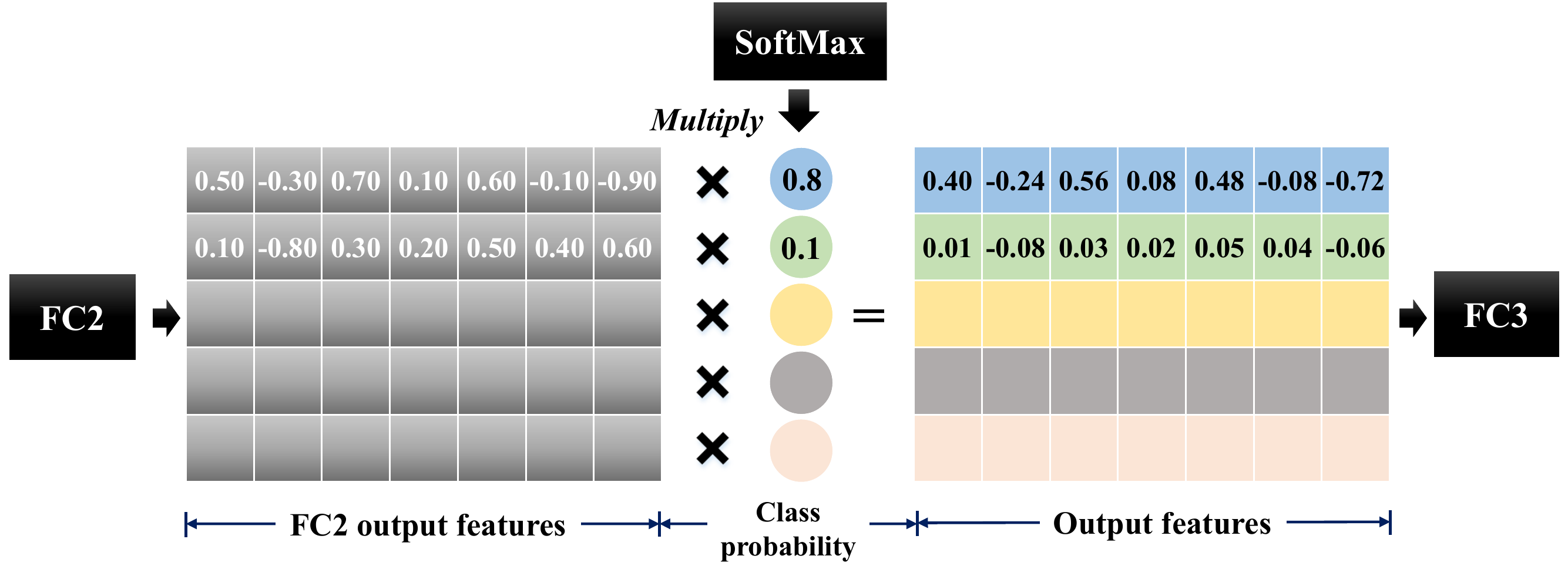} 	
	\caption{Soft Selector for the fusion of classification output and features from FC2. Element-wise multiplication is performed for each row of features (we only show the first two rows here).}\label{fig:selection}
\end{figure}

We fine-tune this entire network over the initialized classification model by minimizing the object function of the joint classification and regression optimization. Note that to enable the classification result indicating which action will begin soon, we set the groundtruth label $z_{t,k}^s = 1$ in the training where $t_{k,start}-T \leq t < t_{k,start}$ for all actions, according to the expected forecast-forward value $T$ as defined in Sec. \ref{sec:problem}. Then, for each frame, the classification output indicates the impending or ongoing action class while the two confidence outputs show the probability to be the start or end point. We set the peak positions of confidences to be the predicted action start (or end) time. Note that since the classification and regression results of the current frame are correlated with the current input and the previous memorized information for the LSTM network, the system does not need to explicitly look back, avoiding sliding window design. 

\end{subsection}

\end{section}

\begin{section}{Experiments}\label{sec:experiments}
In this section, we evaluate the detection and forecast performance of the proposed method on two different skeleton-based datasets. The reason why we choose skeleton-based datasets for experiments is three-fold. First, skeleton joints are captured in the full 3D space, which can provide more comprehensive information for action detection compared to 2D images. Second, the skeleton joints can well represent the posture of human which provide accurate positions and capture human motions. Finally, the dimension of skeleton is low, i.e., 25$\times$3 = 75 values for each frame from Kinect V2. This makes the skeleton based online action detection much attractive for real applications.

\begin{subsection}{Datasets and Settings}
	
Most published RGB-D datasets were generated for the classification task where actions were already pre-segmented \cite{li2010action,Yun2012TwoPerson}. They are only suitable for action recognition. Thus, besides using an existing skeleton-based detection dataset, the Gaming Action Dataset (G3D) \cite{bloom2012g3d}, we collect a new online streaming dataset following  similar rules of previous action recognition datasets, which is much more appropriate for the online action detection problem. In this work, being similar to that in \cite{Zhu2016Co}, the normalization processing on each skeleton frame is performed to be invariant to position.

\textbf{Gaming Action Dataset (G3D).} The G3D dataset contains 20 gaming actions captured by Kinect, which are grouped into seven categories, such as fighting, tennis and golf. Some limitations of this dataset are that the number and occurrence order of actions in the videos are unchanged and the actors are motionless between performing different actions, which make the dataset a little unrealistic.

\textbf{Online Action Detection Dataset (OAD).} This is our newly collected action dataset with long sequences for our online action detection problem. The dataset was captured using the Kinect v2 sensor, which collects color images, depth images and human skeleton joints synchronously. It was captured in a daily-life indoor enviroment. Different actors freely performed 10 actions, including \textit{drinking, eating, writing, opening cupboard, washing hands, opening microwave, sweeping, gargling, throwing trash, and wiping}. We collected 59 long video sequences at 8 fps (in total 103,347 frames of 216 minutes). Note that our low recording frame rate is due to the speed limitation of writing large amount of data (i.e., skeleton, high resolution RGB-D) to the disk of our laptop.

Since the Kinect v2 sensor is capable of providing more accurate depth, our dataset has more accurate tracked skeleton positions compared to previous skeleton datasets. In addition, the acting orders and duration of the actions are arbitrary, which approach the real-life scenarios. The length of each sequence is very long and there are variable idle periods between different actions, which meets the requirements of realistic online action detection from streaming videos.

\textbf{Network and Parameter Settings.} We show the architecture of our network in Fig. \ref{fig:flowchart}. The number of neurons in the deep LSTM network is 100, 100, 110, 110, 100, 100 for the first six layers respectively, including three LSTM layers and three FC layers. The design choice (i.e., LSTM architecture) is motivated by some previous works \cite{du2015hierarchical,Zhu2016Co}. 
The number of neurons in the FC1 layer corresponds to the number of action classes $M+1$ and the number of neurons in the FC2 layer is set to 10$\times$($M+1$). For the FC3 layer, there are two neurons corresponding to the start and end confidences respectively. The forecast response threshold $T$ can be set based on the requirement of the applications. In this paper, we set $T = 10$ (around one second) for the following experiments. The parameter $\sigma$ in (\ref{eq:gauss}) is set to 5. The weight $\lambda$ in the final loss function (\ref{eq:final_loss}) is increased gradually from 0 to 10 during the fine-tuning of the entire network. {Note that we use the same parameter settings for both OAD and G3D datasets.}

For our OAD dataset, we randomly select 30 sequences for training and 20 sequences for testing. The remaining 9 long videos are used for the evaluation of the running speed. For the G3D dataset, we use the same setting as used in \cite{bloom2012g3d}.

\end{subsection}

\begin{subsection}{Action Detection Performance Evaluation}

\textbf{Evaluation criterions.} We use three different evaluation protocols to measure the detection results.
\begin{enumerate}
\item  $F1-$Score. Similar to the protocols used in object detection from images \cite{everingham2010pascal}, we define a criterion to determine a correct detection. A detection is correct when the overlapping ratio $\alpha$ between the predicted action interval $I$ and the groundtruth interval $I^*$ exceeds a threshold, e.g., $60\%$. $\alpha$ is defined as
\begin{equation}
\alpha = \frac{| I \cap I^* |}{| I \cup I^* |},
\end{equation}
where $I \cap I^*$ denotes the intersection of the predicted and groundtruth intervals and $I \cup I^*$ denotes their union. With the above criterion to determine a correction detection, the $F1-$Score is defined as
\begin{equation}
 F1 = 2 \frac{Precision*Recall}{Precision+Recall}.
\end{equation}
\item  $SL-$Score. To evaluate the accuracy of the localization of the start point for an action, we define a Start Localization Score ($SL-$Score) based on the relative distance between the predicted and the groundtruth start time. Suppose that the detector predicts that an action will start at time $t$ and the corresponding groundtruth action interval is $[t_{start}, t_{end}]$, the score is calculated as $e^{-{|t-t_{start}|}/{(t_{end} -t_{start})}}$. For false positive or false negative samples, the score is set to $0$.
\item $EL-$Score. Similarly, the End Localizataion Score ($EL-$Score) is defined based on the relative distance between the predicted and the groundtruth end time.

\end{enumerate}

\textbf{Baselines.} We have implemented serveral baselines for comparison to evaluate the performance of our proposed Joint Classification-Regression RNN model (JCR-RNN),  (i) SVM-SW. We train a SVM detector to detect the action with sliding window design (SW). (ii) RNN-SW. This is based on the baseline method Deep LSTM in \cite{Zhu2016Co}, which employs a deep LSTM network that achieves good results on many skeleton-based action recognition datasets. We train the classifiers and perform the detection based on sliding window design. We set the window size to 10 with step of 5 for both RNN-SW and SVM-SW. We experimentally tried different window sizes and found 10 gives relatively good average performance. (iii) CA-RNN. This is a degenerated version of our model that only consists of the LSTM and classification network, without the regression network involved. We denote it as Classification Alone RNN model (CA-RNN).

\begin{table}[!h] 
	\centering
	\tabcolsep5pt
	\renewcommand\arraystretch{1.2}
	\caption{$F1-$Score on OAD dataset.}\label{tab:osd_detection}
	\begin{tabular}{|c|c|c|c|c|}
		\hline
		\multirow{2}{*}{Actions} & SVM & RNN & CA & JCR \\
		& -SW & -SW \cite{Zhu2016Co} & -RNN & -RNN \\
		\hline
		\hline
		drinking & 0.146	& 0.441	& \textbf{0.584} & 0.574  \\
		\hline
		eating & 0.465 &	0.550 &	\textbf{0.558} &	0.523 \\
		\hline
		writing & 0.645 &	\textbf{0.859} &	0.749 &	0.822 \\
		\hline
		opening cupboard & 0.308 &	0.321 &	0.490 &	\textbf{0.495} \\
		\hline
		washing hands & 0.562 &	0.668 &	0.672 &	\textbf{0.718} \\
		\hline
		opening microwave & 0.607 &	0.665 &	0.468 &	\textbf{0.703} \\
		\hline
		sweeping  & 0.461 &	0.590 &	0.597 &	\textbf{0.643} \\
		\hline
		gargling & 0.437 &	0.550 &	0.579 &	\textbf{0.623} \\
		\hline
		throwing trash & 0.554 &	\textbf{0.674} &	0.430 &	0.459 \\
		\hline
		wiping  & \textbf{0.857}	 & 0.747 &	0.761 &	0.780 \\
		\hline
		\textbf{average} & 0.540	 & 0.600 &	0.596 &	\textbf{0.653} \\
		\hline
	\end{tabular}
\end{table}

\textbf{Dectection Performance.} Table \ref{tab:osd_detection} shows the $F1-$Score of each action class and the average $F1-$Score of all actions on our OAD Dataset. From Table 1, we have the following observations. (i) The RNN-SW method achieves 6\% higher $F1-$Score than the SVM-SW method. This demonstrates that RNNs can better model the temporal
dynamics. (ii) Our JCR-RNN outperforms the RNN-SW method by 5.3\%. Despite RNN-SW, CA-RNN and JCR-RNN methods all use RNNs for feature learning, one difference is that our schemes are end-to-end trainable without the involvement of sliding window. Therefore, the improvements clearly demonstrate that our end-to-end schemes are more efficient than the classical sliding window scheme. (iii) Our JCR-RNN further improves over the CA-RNN and achieves the best performance. It can be seen that incorporating the regression task into the network and jointly optimizing classification-regression make the localization more accurate and enhance the detection accuracy.

To further evaluate the localization accuracy, we calcuate the $SL-$ and $EL-$Scores on the Online Action Dataset. The average scores of all actions are shown in Table \ref{tab:osd_loc}. We can see the proposed scheme achieves the best localization accuracy.

\begin{table}[t] 
	\centering
	\tabcolsep5pt
	\renewcommand\arraystretch{1.2}
	\caption{$SL-$ and $EL-$Score on the OAD dataset.}\label{tab:osd_loc}
	\begin{tabular}{|c|c|c|c|c|}
		\hline
		\multirow{2}{*}{Scores} &  SVM & RNN & CA & JCR \\
		& -SW & -SW \cite{Zhu2016Co} & -RNN & -RNN \\
		\hline
		\hline
		$SL-$  & 0.316 & 	0.366 & 	0.378 & 	\textbf{0.418} \\
		\hline
		$EL-$ & 0.325 & 	0.376 & 	0.382 & 	\textbf{0.443} \\
		\hline
	\end{tabular}
\end{table}

For the G3D dataset, we evaluate the performance in terms of the three types of scores for the seven categories of sequences. To save space, we only show the results for the first two categories \textit{Fighting} and \textit{Golf} in Table \ref{tab:g3d_F1score} and \ref{tab:g3d_score_SL_EL}, and more results which have the similar trends can be found in the supplementary material. The results are consistent with the experiments on our own dataset. We also compare these methods using the evaluation metric action-based $F1$ as defined in \cite{bloom2012g3d}, which treats the detection of an action as correct when the predicted start point is within 4 frames of the groundtruth start point for that action. Note that the action-based $F1$ only considers the accuracy of the start point. The results are shown in Table \ref{tab:g3d_score1}. The method in \cite{bloom2012g3d} uses a traditional boosting algorithm \cite{freund1996experiments} and its scores are significantly lower than other methods.

\begin{table}[t]
\centering
\makebox[0pt][c]{\parbox{0.95\textwidth}{%
        \begin{minipage}[b]{0.45\hsize}\centering
  \small
  \tabcolsep1pt
		\caption{$SL-$ and $EL-$Score on the G3D Dataset.}\label{tab:g3d_score_SL_EL}
		\begin{tabular}{|c|c|c|c|c|c|}
			\hline
			Action & \multirow{2}{*}{Scores} & SVM & RNN & CA & JCR \\
			Category &   & -SW & -SW \cite{Zhu2016Co} & -RNN & -RNN \\
			\hline
			\hline
			\multirow{2}{*}{Fighting}
			& $SL-$ & 0.318 & 0.412 & 0.512 & \textbf{0.528} \\
			\cline{2-6}
			& $EL-$ & 0.328 & 0.419 & 0.525 & \textbf{0.557} \\
			\hline
			
			\multirow{2}{*}{Golf} & $SL-$ & 0.553 & 0.635 & 0.789 & \textbf{0.793} \\
			\cline{2-6}
			& $EL-$ & 0.524 & 0.656 & 0.791 & \textbf{0.836} \\
			\hline
		\end{tabular}
    \end{minipage}
   \hfill
    \begin{minipage}[b]{0.45\hsize}\centering
  \small
  \tabcolsep1pt
  \caption{$F1-$Score on G3D.}\label{tab:g3d_F1score}
		\begin{tabular}{|c|c|c|c|c|}
			\hline
			Action & SVM & RNN & CA & JCR \\
			Category   & -SW & -SW \cite{Zhu2016Co} & -RNN & -RNN \\
			\hline
			\hline
			Fighting  & 0.486 & 0.613 & 0.700 & \textbf{0.735} \\
			\hline
			Golf &  0.680 & 0.745 & 0.900 & \textbf{0.967} \\
			\hline
		\end{tabular}
    \end{minipage}
}}
\end{table}

			

\begin{table}[t]
\centering
\makebox[0pt][c]{\parbox{0.95\textwidth}{%
        \begin{minipage}[b]{0.5\hsize}\centering
  \small
  \tabcolsep1pt
		 \caption{Action-based $F1$ \cite{bloom2012g3d} on G3D.}\label{tab:g3d_score1}
   \begin{tabular}{|c|c|c|c|c|c|}
  \hline
Action & G3D  & SVM & RNN & CA & JCR \\
Category   & \cite{bloom2012g3d} & -SW & -SW \cite{Zhu2016Co} & -RNN & -RNN \\
\hline
\hline
 Fighting  & 58.54 & 76.72 & 83.28 & 94.00 & \textbf{96.18} \\
\hline
 Golf & 11.88 & 45.00 & 55.00 & 50.00 & \textbf{70.00} \\
\hline
  \end{tabular}
    \end{minipage}
   \hfill
    \begin{minipage}[b]{0.45\hsize}\centering
  \small
  \tabcolsep1pt
  \renewcommand\arraystretch{1.2}
\caption{Average running time (seconds per sequence).}\label{tab:running_time}
\begin{tabular}{|c|c|c|}
\hline
SVM-SW & RNN-SW \cite{Zhu2016Co} & JCR-RNN \\
\hline
\hline
1.05  & 3.14 & 2.60 \\
\hline
\end{tabular}
    \end{minipage}
}}
\end{table}

\end{subsection}

\begin{subsection}{Action Forecast Performance Evaluation}
\begin{figure*}[t] 
\begin{center}
\subfigure[Forecast of start.]{
 \centering\includegraphics[width=0.4\linewidth]{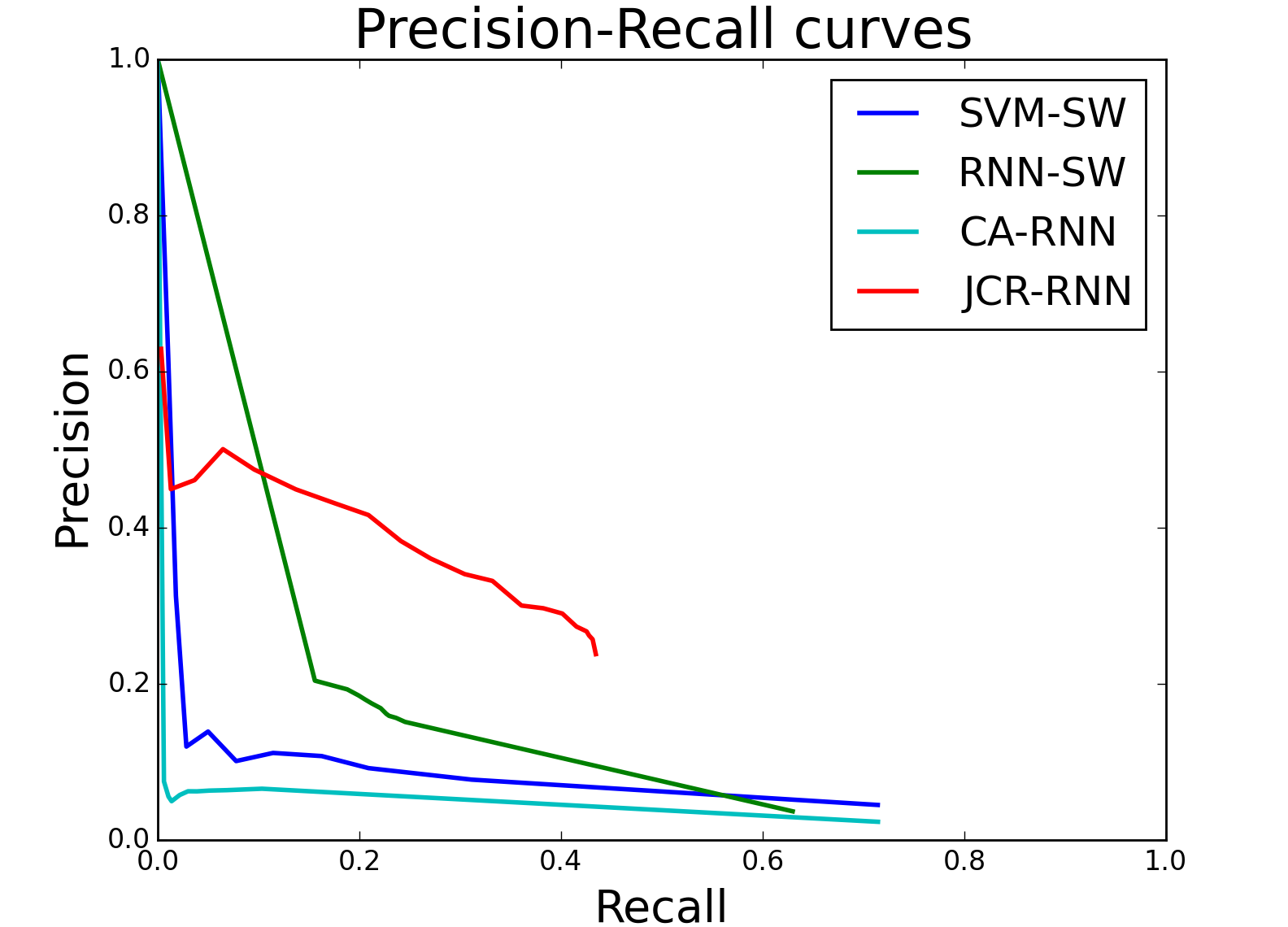} 	
}
\hspace{6mm}
\subfigure[Forecast of end.]{
  \centering\includegraphics[width=0.4\linewidth]{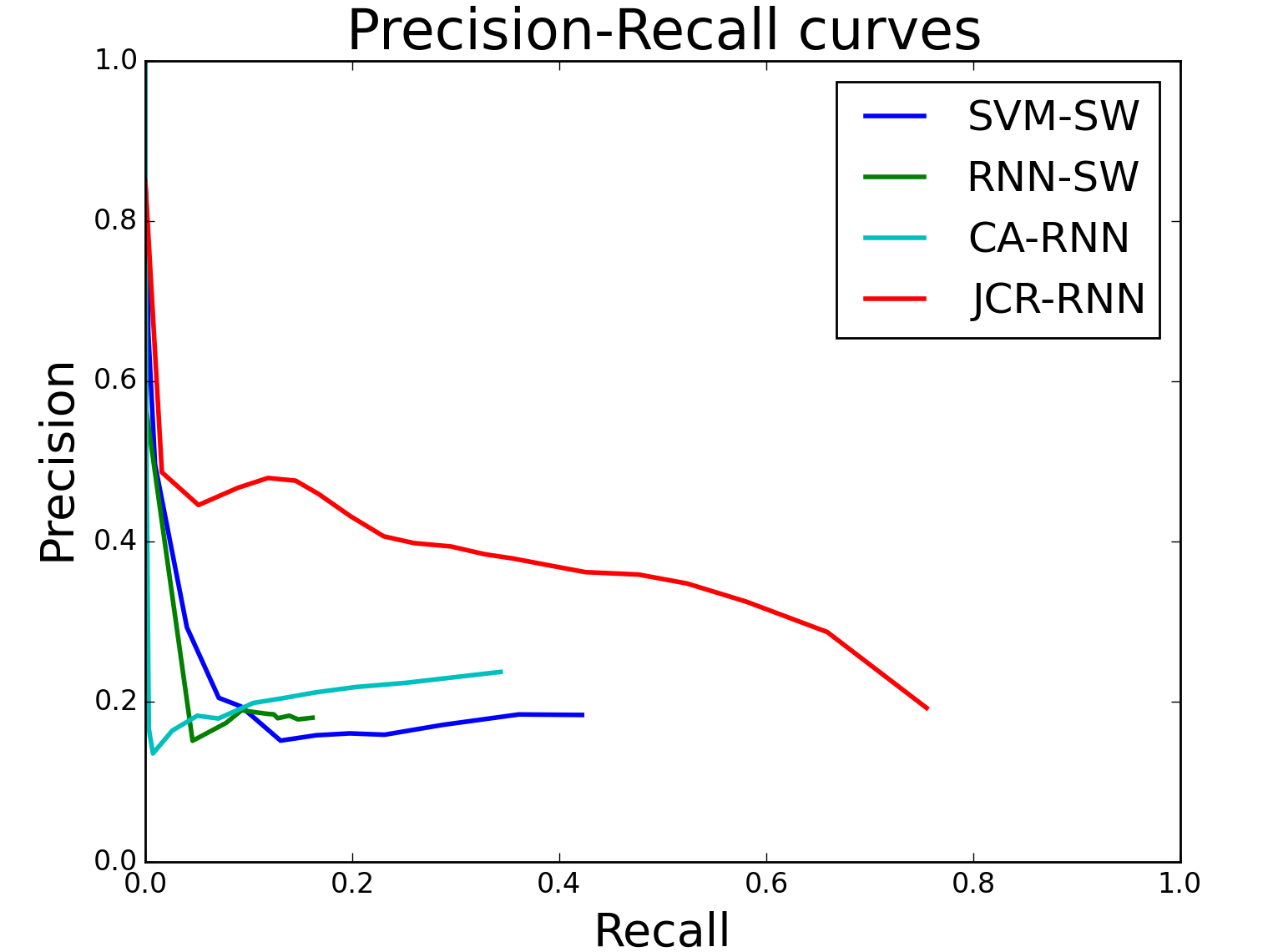} 	
}
\caption{The Precision-Recall curves of the start and end time forecast with different methods on the OAD dataset. Overall JCR-RNN outperforms other baselines by a large margin. This figure is best seen in color. }
\label{fig:pr_curve}
\end{center}
\end{figure*}
\textbf{Evaluation criterion.} As explained in Section \ref{sec:problem}, the system is expected to forecast whether the action will start or end within $T$ frames prior to its occurrence. To be considered as a true positive start forecast, the forecast should not only predict the impending action class, but also do so within a reasonable interval, i.e., $[t_{start} - T, t_{start}]$ for an action starting at $t_{start}$. This rule is also applied to end forecast. We use the Precision-Recall Curve to evaluate the performance of the action forecast methods. Note that both precision and recall are calculated on the frame-level for all frames.

\textbf{Baselines.} Since there is no previous method proposed for the action forecast problem, we use a simple strategy to do the forecast based on the above detection baseline methods. For SVM-SW, RNN-SW, CA-RNN, they will output the probability $q_{t,j}$ for each action class $j$ at each time step $t$. At time $t$, when the probability $q_{t,j}$ of action class $j$ is larger than a predefined threshold $\beta_s$, we consider that the action of class $j$ will start soon. Similarly, during an ongoing period of the action of class $j$, when the probability $q_{t,j}$ is smaller than another threshold $\beta_e$, we consider this action to end soon.

\textbf{Forecast Performance.} The peak point of the regressed confidence curve is considered as the start/end point in the test. When the current confidence value is higher than $\theta_s$ but ahead of the peak, this frame forecasts that the action will start soon. By adjusting the confidence thresholds $\theta_s$ and $\theta_e$ in our method, we draw the Precision-Recall curves for our method. Similiarly, we draw the curves for the baselines by adjusting $\beta_s$ and $\beta_e$. We show them in Fig. \ref{fig:pr_curve}. The performance of the baselines is significantly inferior to  our method JCR-RNN. This suggests that only using the traditional detection probability is not suitable for forecasting. One important reason is that the frames before the start time are simply treated as background samples in the baselines but actually they contain evidences. While in our regression task, we deal with these frames using different confidence values to guide the network to explore the hidden starting or ending patterns of actions. In addition, we note that the forecast precision of all the methods are not very high even though our method is much better, e.g., precision is $28\%$ for start forecast and $37\%$ for end forecast when recall is $40\%$. This is because the forecast problem itself is a difficult problem. For example, when a person is writing on the board, it is difficult to forecast whether he will finish writing soon.  

Fig. \ref{fig:cm} shows the confusion matrix of the start forecast by our proposed method. This confusion matrix represents the relationships between the predicted start action class and the groundtruth class. The shown confusion matrix is obtained when the recall rate equals to 40\%. From this matrix, although there are some missed or wrong forecasts, most of the forecasts are correct. In addition, there are a few interesting observations. For example, the action \textit{eating} and \textit{drinking} may have similar poses before they start. Action \textit{gargling} and \textit{washing hands} are also easy to be mixed up when forecasting since the two actions both need to turn on the tap before starting. Taking into account human-object interaction should help reduce the ambiguity and we will leave it for future work.

\begin{figure}[t]
		\centering\includegraphics[width=0.5\linewidth]{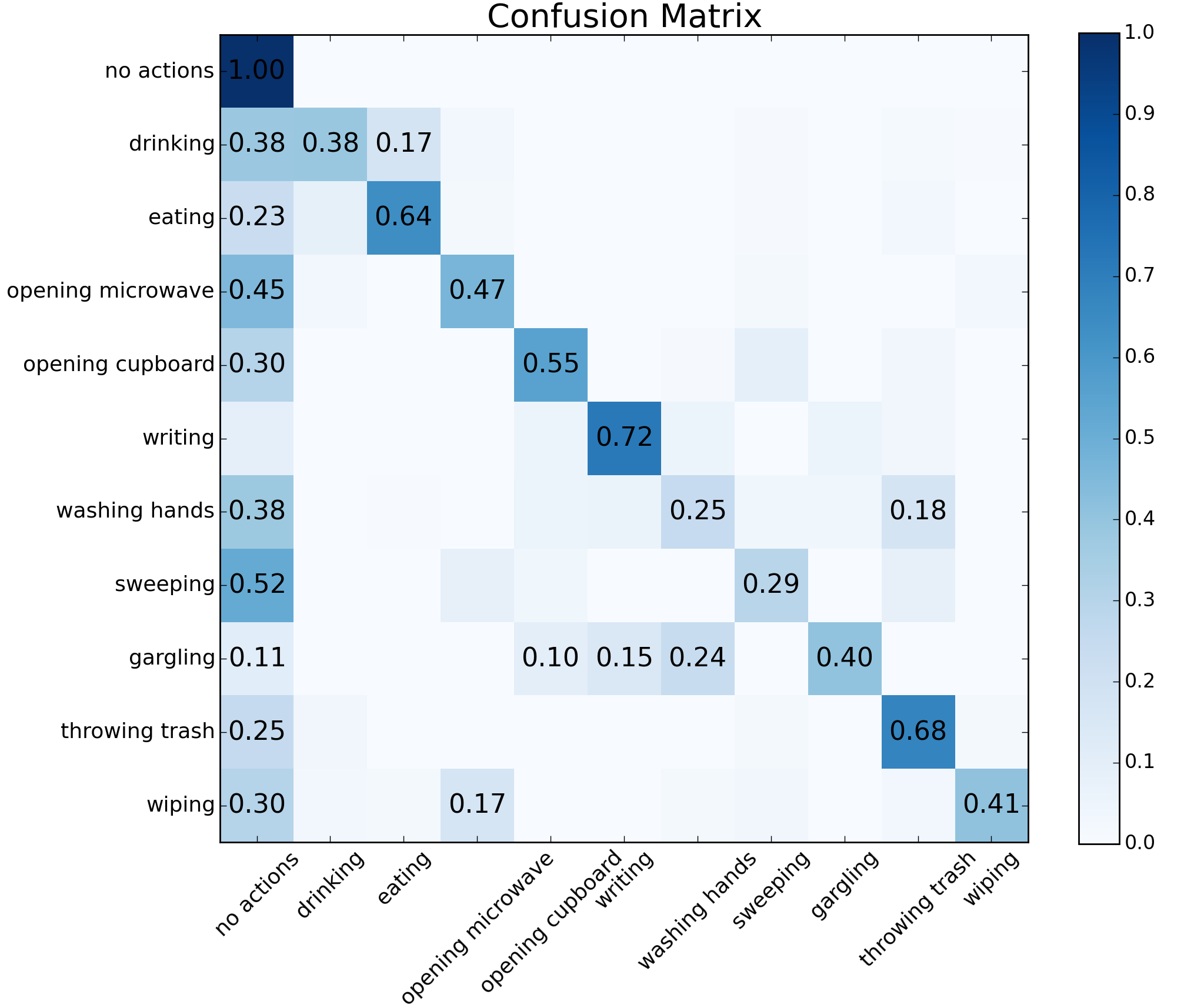}
		\caption{Confusion Matrix of start forecast on the OAD dataset. Vertical axis: groundtruth class; Horizontal axis: predicted class.}
		\label{fig:cm}
\end{figure}

\end{subsection}
\begin{subsection}{Comparison of Running Speeds}
In this section, we compare the running speeds of different methods. Table \ref{tab:running_time} shows the average running time on 9 long sequences, which has 3200 frames on average. SVM-SW has the fastest speed because of its small model compared with the deep learning methods. The RNN-SW runs slower than our methods due to its sliding window design. We can notice the running speed for the action detection based on skeleton input is rather fast, being 1230 fps for the JCR-RNN approach. This is because the dimension of skeleton is low (25$\times$3 = 75 values for each frame) in comparison with RGB input. This makes the skeleton based online action detection much attractive for real applications.


\end{subsection}

\end{section}

\begin{section}{Conclusion and Future Work}\label{sec:conclusion}
In this paper, we propose an end-to-end Joint Classification-Regression RNN to explore the action type and better localize the start and end points on the fly.  We leverage the merits of the deep LSTM network to capture the complex long-range temporal dynamics and avoid the typical sliding window design. We first pretrain the classification network for the frame-wise action classification. Then with the incorporation of the regression network, our joint model is capable of not only localizing the start and end time of actions more accurately but also forecasting their occurrence in advance. Experiments on two datasets demonstrate the effectiveness of our method. In the future work, we will introduce more features, such as appearance and human-object interaction information, into our model to further improve the detection and forecast performance.
\\

\end{section}

\noindent\textbf{Acknowledgement.} This work was supported by National High-tech Technology R$\&$D Program (863 Program) of China under Grant 2014AA015205, National Natural Science Foundation of China under contract No.61472011 and No.61303178, and Beijing Natural Science Foundation under contract No.4142021.

\clearpage

\bibliographystyle{splncs}
\bibliography{egbib}

\clearpage

\textbf{\Large{Supplementary}}

\setcounter{section}{0}

\begin{section}{Experimental Results on the OAD Dataset}

\begin{subsection}{Evaluation of the Soft Selector Module}
In this section, we evaluate the influence of the Soft Selector Module, which described in Section 4.3. We implement a variant version of our method by removing the Soft Selector module and directly linking the FC2 and FC3 layers. Table \ref{tab:softselector} and Table \ref{tab:slel_softselect} compare the $F1-$Score, $SL-$ and $EL-$Scores of our method with and without soft selctor on the OAD Dataset. Note that we use the same parameters for the two settings. We can see that the incorporation of the Soft Selector can bring significant improvement.

\begin{table}[hbtp] 
	\centering
	\tabcolsep5pt
	\renewcommand\arraystretch{1.2}
	\caption{$F1-$Score on OAD dataset.}\label{tab:softselector}
	\begin{tabular}{|c|c|c|}
		\hline
		Actions & w/o Soft Selector & with Soft Selector \\
		\hline
		\hline
		drinking & 0.445 & \textbf{0.574}  \\
		\hline
		eating & \textbf{0.542} &	0.523 \\
		\hline
		writing & 	0.714 &	\textbf{0.822} \\
		\hline
		opening cupboard & \textbf{0.526} &	0.495 \\
		\hline
		washing hands & 0.688 &	\textbf{0.718} \\
		\hline
		opening microwave & 0.697 &	\textbf{0.703} \\
		\hline
		sweeping  & 0.547 &	\textbf{0.643} \\
		\hline
		gargling & 0.478 &	\textbf{0.623} \\
		\hline
		throwing trash & \textbf{0.608} &	0.459 \\
		\hline
		wiping  & 0.691 &	\textbf{0.780} \\
		\hline
		overall & 0.616 &	\textbf{0.653} \\
		\hline
	\end{tabular}
\end{table}

\begin{table}[hbtp] 
	\centering
	\tabcolsep5pt
	\renewcommand\arraystretch{1.2}
	\caption{$SL-$ and $EL-$Scores on the OAD dataset.}\label{tab:slel_softselect}
	\begin{tabular}{|c|c|c|}
		\hline
		Scores & w/o Soft Selector & with Soft Selector \\
		\hline
		\hline
		$SL-$  & 0.389 & 	\textbf{0.418} \\
		\hline
		$EL-$ & 0.413 & 	\textbf{0.443} \\
		\hline
	\end{tabular}
\end{table}

\end{subsection}


\begin{subsection}{Sliding Window Size of the Baselines}
Since the baseline methods SVM-SW and RNN-SW \cite{Zhu2016Co} are based on the sliding window schemes, we report the $F1-$Score of these methods using different window size (number of frames) in Table \ref{tab:SVM_windowsize} and \ref{tab:RNN_windowsize}. The correponding $SL-$ and $EL-$ scores are shown in Table \ref{tab:SVM_windowsize_SE} and \ref{tab:RNN_windowsize_SE}. In the experiments, the stride of the window is set to be half of the window size.

\begin{table}[hbtp] 
	\centering
	\tabcolsep5pt
	\renewcommand\arraystretch{1.2}
	\caption{$F1-$Score of using different window sizes (ws) for SVM-SW method on OAD dataset.}\label{tab:SVM_windowsize}
	\begin{tabular}{|c|c|c|c|c|c|}
		\hline
		\multirow{2}{*}{Actions} & \multicolumn{4}{|c|}{SVM-SW} & \multirow{2}{*}{JCR-RNN} \\
		\cline{2-5}
		& ws=5 & ws=10 & ws=20 & ws=40 &  \\
		\hline
		\hline
		drinking & 0.291 &	0.146&	0.000&	0.000 & \textbf{0.574}  \\
		\hline
		eating & 0.507&	0.465	&\textbf{0.548}	&0.050 &	0.523 \\
		\hline
		writing & 	0.671 &	0.645&	0.792&	0.542 &	\textbf{0.822} \\
		\hline
		opening cupboard & 0.284&	0.308&	0.352&	0.033 &	\textbf{0.495} \\
		\hline
		washing hands & 0.501&	0.562	&0.650&	0.308 &	\textbf{0.718} \\
		\hline
		opening microwave &0.521&	0.607	&0.590&	0.492 &	\textbf{0.703} \\
		\hline
		sweeping  & 0.434&	0.461&	0.515&	0.498 &	\textbf{0.643} \\
		\hline
		gargling & 0.466&	0.437&	0.433&	0.083 &	\textbf{0.623} \\
		\hline
		throwing trash & 0.475&	0.554&	0.315&	0.000 &	\textbf{0.459} \\
		\hline
		wiping  & \textbf{0.867}&	0.857&	0.748&	0.433 &	0.780 \\
		\hline
		overall & 0.525&	0.540&	0.565&	0.334 &	\textbf{0.653} \\
		\hline
	\end{tabular}
\end{table}

\begin{table}[hbtp] 
	\centering
	\tabcolsep5pt
	\renewcommand\arraystretch{1.2}
	\caption{$F1-$Score of using different window sizes (ws) for RNN-SW \cite{Zhu2016Co} method on OAD dataset.}\label{tab:RNN_windowsize}
	\begin{tabular}{|c|c|c|c|c|c|}
		\hline
		\multirow{2}{*}{Actions} & \multicolumn{4}{|c|}{RNN-SW \cite{Zhu2016Co}} & \multirow{2}{*}{JCR-RNN} \\
		\cline{2-5}
		& ws=5 & ws=10 & ws=20 & ws=40 &  \\
		\hline
		\hline
		drinking & 0.495&	0.441&	0.142	&0.033 & \textbf{0.574}  \\
		\hline
		eating & \textbf{0.525}&	0.550	&0.532	&0.183 &	0.523 \\
		\hline
		writing & 	0.665	&\textbf{0.859}&	0.837&	0.725 &	0.822 \\
		\hline
		opening cupboard & 0.331&	0.321	&0.413	&0.217 &	\textbf{0.495} \\
		\hline
		washing hands & 0.588	&0.668&	\textbf{0.867}&	0.517 &	0.718 \\
		\hline
		opening microwave &0.628	&0.665&	0.680&	0.647 &	\textbf{0.703} \\
		\hline
		sweeping  & 0.475&	0.590&	\textbf{0.811}	&0.783 &	0.643 \\
		\hline
		gargling & 0.426&	0.550&	0.534&	0.108 &	\textbf{0.623} \\
		\hline
		throwing trash & 0.434&	\textbf{0.674}	&0.325&	0.050 &	0.459 \\
		\hline
		wiping  & 0.734	&0.747&	\textbf{0.797}&	0.550 &	0.780 \\
		\hline
		overall & 0.512	&0.600	&0.627	&0.476 &	\textbf{0.653} \\
		\hline
	\end{tabular}
\end{table}

\begin{table}[hbtp] 
	\centering
	\tabcolsep5pt
	\renewcommand\arraystretch{1.2}
	\caption{$SL-$ and $EL-$Scores of using different window sizes (ws) for SVM-SW method on OAD dataset.}\label{tab:SVM_windowsize_SE}
	\begin{tabular}{|c|c|c|c|c|c|}
		\hline
		\multirow{2}{*}{Scores} & \multicolumn{4}{|c|}{SVM-SW} & \multirow{2}{*}{JCR-RNN} \\
		\cline{2-5}
		& ws=5 & ws=10 & ws=20 & ws=40 &  \\
		\hline
		\hline
		$SL-$ & 0.288&	0.316&	0.339&	0.182 & \textbf{0.418}  \\
		\hline
		$EL-$ & 0.300&	0.325&	0.343	&0.184 & \textbf{0.443} \\
		\hline
	\end{tabular}
\end{table}

\begin{table}[hbtp] 
	\centering
	\tabcolsep5pt
	\renewcommand\arraystretch{1.2}
	\caption{$SL-$ and $EL-$Scores of using different window sizes (ws) for RNN-SW \cite{Zhu2016Co} method on OAD dataset.}\label{tab:RNN_windowsize_SE}
	\begin{tabular}{|c|c|c|c|c|c|}
		\hline
		\multirow{2}{*}{Scores} & \multicolumn{4}{|c|}{RNN-SW \cite{Zhu2016Co}} & \multirow{2}{*}{JCR-RNN} \\
		\cline{2-5}
		& ws=5 & ws=10 & ws=20 & ws=40 &  \\
		\hline
		\hline
		$SL-$ & 0.287&	0.366	&0.393	&0.276 & \textbf{0.418}  \\
		\hline
		$EL-$ & 0.291&	0.376&	0.401&	0.274& \textbf{0.443} \\
		\hline
	\end{tabular}
\end{table}

\end{subsection}

\end{section}

\begin{section}{Experimental Results on the G3D Dataset}

In this section, we show experimental results on all the seven categories of videos in the G3D dataset \cite{bloom2012g3d}. Note that we have only shown the results on the first two categories of videos in the paper due to the space limitation. We evaluate the performance in terms of the $F1-$Score and $SL-$ and $EL-$Score as described in this paper, and summarize them in Table \ref{tab:g3d_F1score1} and Table \ref{tab:g3d_score_SL_EL1}. Table \ref{tab:g3d_score11} shows the comparsion of these methods using the evaluation metric of action-based $F1$ as defined in \cite{bloom2012g3d}.

\begin{table}[hbtp]
\centering
  \tabcolsep5pt
  \renewcommand\arraystretch{1.2}
		\caption{$F1-$Score on the G3D Dataset.}\label{tab:g3d_F1score1}
		\begin{tabular}{|c|c|c|c|c|}
			\hline
			Action Category & SVM-SW & RNN-SW \cite{Zhu2016Co} & CA-RNN & JCR-RNN \\
			\hline
			\hline
			Fighting  & 0.486 & 0.613 & 0.700 & \textbf{0.735} \\
			\hline
			Golf &  0.680 & 0.745 & 0.900 & \textbf{0.967} \\
			\hline
			Tennis &  0.598 & 0.480 & 0.774 & \textbf{0.788} \\
			\hline
			Bowling &  0.667 & 0.889 & 1.000 & \textbf{1.000} \\
			\hline
			FPS &  0.571 & \textbf{0.581} & 0.378 & 0.523 \\
			\hline
			Driving &  1.000 & 1.000 & 1.000 & \textbf{1.000} \\
			\hline
			Misc &  0.712 & 0.742 & 0.813 & \textbf{0.862} \\
			\hline

		\end{tabular}
\end{table}

\begin{table}[hbtp]
\centering
  \tabcolsep5pt
  \renewcommand
  \arraystretch{1.2}
		\caption{$SL-$ and $EL-$Score on the G3D Dataset.}\label{tab:g3d_score_SL_EL1}
		\begin{tabular}{|c|c|c|c|c|c|}
			\hline
			Action Category & {Scores} & SVM-SW & RNN-SW \cite{Zhu2016Co} & CA-RNN & JCR-RNN \\
			\hline
			\hline
			\multirow{2}{*}{Fighting}
			& $SL-$ & 0.318 & 0.412 & 0.512 & \textbf{0.528} \\
			& $EL-$ & 0.328 & 0.419 & 0.525 & \textbf{0.557} \\
			\hline
			
			\multirow{2}{*}{Golf} 
			& $SL-$ & 0.553 & 0.635 & 0.789 & \textbf{0.793} \\
			& $EL-$ & 0.524 & 0.656 & 0.791 & \textbf{0.836} \\
			\hline

			\multirow{2}{*}{Tennis} 
			& $SL-$ & 0.444 & 0.338 & 0.605 & \textbf{0.665} \\
			& $EL-$ & 0.460 & 0.333 & 0.617 & \textbf{0.667} \\
			\hline

			\multirow{2}{*}{Bowling} 
			& $SL-$ & 0.612 & 0.777 & 0.933 & \textbf{0.959} \\
			& $EL-$ & 0.550 & 0.713 & 0.816 & \textbf{0.861} \\
			\hline

			\multirow{2}{*}{FPS} 
			& $SL-$ & 0.351 & \textbf{0.388} & 0.183 & 0.311\\
			& $EL-$ & 0.353 & \textbf{0.393} & 0.199 & 0.327 \\
			\hline

			\multirow{2}{*}{Driving} 
			& $SL-$ & \textbf{0.991} & 0.983 & 0.957 & 0.955 \\
			& $EL-$ & 0.975 & 0.975 & 0.964 & \textbf{0.975} \\
			\hline

			\multirow{2}{*}{Misc} 
			& $SL-$ & 0.487 & 0.593 & 0.609 & \textbf{0.614} \\
			& $EL-$ & 0.515 & 0.612 & 0.690 & \textbf{0.766} \\
			\hline
		\end{tabular}
\end{table}

\begin{table}[hbtp] 
\centering
  \tabcolsep5pt
  \renewcommand\arraystretch{1.2}
   \caption{Action-based $F1$ \cite{bloom2012g3d} on the G3D Dataset.}\label{tab:g3d_score11}
   \begin{tabular}{|c|c|c|c|c|c|}
  \hline
Action Category & G3D \cite{bloom2012g3d} & SVM-SW & RNN-SW \cite{Zhu2016Co} & CA-RNN & JCR-RNN \\
\hline
\hline

 Fighting  & 58.54 & 76.72 & 83.28 & 94.00 & \textbf{96.18} \\
\hline
 Golf & 11.88 & 45.00 & 55.00 & 50.00 & \textbf{70.00} \\
\hline
 Tennis & 14.85 & 37.57 & 36.68 & 59.04 & \textbf{62.38} \\
\hline
 Bowling & 31.58 & 22.22 & 44.44 & 44.44 & \textbf{66.67} \\
\hline
 FPS & 13.65 & 35.35 & \textbf{39.89} & 23.69 & 33.85 \\
\hline
 Driving & 2.5 & 39.99 & 50.00 & 19.99 & \textbf{50.00} \\
\hline
 Misc & 18.13 & 53.32 & 65.24 & 73.81 & \textbf{86.19} \\
\hline
  \end{tabular}
\end{table}

\end{section}




\end{document}